\documentclass[final,5p,times,twocolumn,authoryear]{elsarticle}

\usepackage{amssymb}
\usepackage{lipsum}
\usepackage{amsmath}
\setcitestyle{numbers,square}
\usepackage{tabularx}
\usepackage{makecell}

\journal{High Energy Astrophysics}

\begin{document}

\begin{frontmatter}



\title{Beyond Binary Classification: Multiclass Machine Learning for Diabetes and Prediabetes Risk Prediction Using NHANES}


\author[first]{Aman Mujahid}

\author[second]{Anita Penkova}

\affiliation[first]{organization={Thomas Lord Department of Computer Science, Viterbi School of Engineering, University of Southern California},
            city={Los Angeles},
            postcode={90089}, 
            state={CA},
            country={USA}}

\affiliation[second]{organization={Department of Aerospace and Mechanical Engineering, Viterbi School of Engineering, University of Southern California},
            city={Los Angeles},
            postcode={90089}, 
            state={CA},
            country={USA}}

\begin{abstract}

Detecting diabetes risk early is important for preventing disease progression. Most existing machine learning approaches focus on binary classification, which limits their ability to capture intermediate risk states such as prediabetes. We develop a supervised learning framework to classify adults as normal, prediabetic, or diabetic using National Health and Nutrition Examination Survey data. The framework uses routinely collected demographic, clinical, and lifestyle variables from 15,356 adults, with glycated hemoglobin (HbA1c) used only to define labels and excluded from input features to prevent leakage. We compare conventional machine learning and transformer-based models and examine two binary formulations: normal versus prediabetes, and normal versus combined prediabetes and diabetes. Models are evaluated on a held-out test set using precision, sensitivity, specificity, and F1 score, with 5-fold cross-validation and 95\% confidence intervals assessing stability. In the multiclass setting, performance is broadly consistent across models, with accuracy up to 0.72 and macro-averaged F1 of 0.70. Performance is stronger for the normal and diabetes categories, with normal-class F1 reaching 0.79 and diabetes sensitivity reaching 0.80, and XGBoost achieving a diabetes precision of 0.73 and specificity of 0.96. Prediabetes remains the most challenging class, with F1 scores of 0.54 to 0.56 for conventional models and 0.53 to 0.55 for transformers, reflecting its overlap with both neighboring states. For this class, XGBoost achieves a precision of 0.49, sensitivity of 0.65, and specificity of 0.74. When the task is reduced to normal versus prediabetes, F1 scores rise modestly to 0.62 to 0.64. A more substantial improvement is observed under a broader screening formulation combining prediabetes and diabetes, with F1 scores of 0.73 to 0.74 across all models and XGBoost reaching a sensitivity of 0.78 and specificity of 0.77. These findings suggest that feature overlap along the biological continuum between normal and diabetic states contributes more to the difficulty of prediabetes detection than model architecture. Multiclass prediction provides a more clinically informative representation of metabolic risk progression but highlights a trade-off between clinically meaningful stratification and predictive accuracy.

\end{abstract}




\end{frontmatter}

\newpage




\section{Introduction}
\label{introduction}

Diabetes mellitus remains a major public health challenge worldwide, with prevalence continuing to increase due to demographic shifts and behavioral risk factors. Although long-term complications of diabetes, which include cardiovascular disease, kidney dysfunction, and vision impairment, are well documented, the condition often advances quietly over time. A lot of individuals pass through a prediabetic phase marked by impaired glucose regulation and few or no symptoms. As a result, many people who are already at elevated risk remain unaware of their condition to the point where it reaches an advanced stage. Detecting abnormal patterns in blood glucose earlier can make a big difference by enabling timely preventive interventions.

Prediabetes represents a state of transition between normal glucose control and diabetes, where early changes in behavior and daily habits can reduce the risk of progression. Evidence from the Diabetes Prevention Program demonstrated that both lifestyle intervention and metformin reduced the incidence of type 2 diabetes among individuals at elevated risk, with lifestyle intervention producing the greater reduction \cite{knowler2002reduction}. Despite its clinical importance, this condition is frequently overlooked since routine screening practices often merge it with normal glycemic status. Therefore, many predictive models frame diabetes risk as a binary problem, differentiating only the presence and absence of disease. Keeping prediabetes separate, as in treating it as a distinct category, paints a clearer picture of the risk and opens the door to faster treatment.

Glycemic status can be assessed using several commonly collected biomarkers such as fasting plasma glucose, oral glucose tolerance tests, and glycated hemoglobin (HbA1c). Clinical guidelines define clear HbA1c ranges corresponding to normal glycemic control, prediabetes, and diabetes \cite{ada2026}, which makes it a practical tool for classifying individuals within a large population. For models designed to assess risk before a clinical diagnosis is made, however, including HbA1c as a predictor can leak information that directly defines the outcome, making performance look stronger than it would in a real screening setting.

Machine learning methods have been used for quite a while to assess diabetes risk by detecting patterns across a wide range of demographic, clinical, and behavioral information. Compared with conventional statistical models, these methods can better capture complex and nonlinear associations in observational health data. Although there have been great advances in machine learning for the field of diabetes prediction, most studies that have been done so far rely on binary classification, limiting their ability to model intermediate metabolic states like prediabetes.

To fill these gaps, we used data from the United States National Health and Nutrition Examination Survey (NHANES), a large observational dataset that includes demographic, clinical, and lifestyle information from adults across the general population. Using this data, we developed a supervised learning framework that classifies individuals as normal, prediabetic, or diabetic based on established HbA1c criteria, while excluding HbA1c from the input features to prevent label leakage. We evaluated multiple gradient boosting models, a soft voting ensemble, and three transformer-based architectures, and reported macro-averaged metrics to focus on balanced performance across classes. We also ran an exploratory subgroup analysis with fairness-aware reweighting to check how consistently prediabetes is detected among the demographic groups. By comparing multiclass and binary screening tasks, we tested whether the main challenge comes from model capacity or from overlap between adjacent glycemic states. This work makes the following contributions:

\begin{itemize}

\item We show that prediabetes remains difficult to separate from adjacent glycemic states across both conventional machine learning models and transformer-based architectures, suggesting that a substantial portion of the limitation arises from feature overlap within the metabolic continuum rather than insufficient model complexity.

\item We show that prediction performance is influenced more strongly by problem formulation than by model selection, with substantial gains observed when prediabetes and diabetes are merged into a broader abnormal glycemic category.

\item We distinguish realistic screening-oriented prediction from diagnostic reconstruction by excluding HbA1c from the predictor set, allowing model performance to reflect information available prior to formal diagnosis.

\end{itemize}

\section{Related Work}
\label{related_work}

\subsection{Machine Learning Approaches for Diabetes Risk Prediction}

Much of the machine learning work on diabetes has focused on finding algorithms that can improve prediction accuracy and better separate individuals with diabetes from those without it. For instance, Xu and Wang combined random forest based feature selection with XGBoost and showed that this approach performed better than several standard classification methods on the Pima Indians Diabetes dataset \cite{xu2019rf_xgboost}. Deberneh and Kim used electronic health records to predict whether patients would be normal, prediabetic, or diabetic in the following year, and reported that ensemble models performed better than individual classifiers \cite{healthcare_xgboost_2020}. More recent studies have also used large population-level datasets such as NHANES, where XGBoost achieved the highest AUC among several compared classifiers \cite{riverosperez2025nhanes, yang2026nhanes_highrisk}, and others have combined several models through ensemble and stacking frameworks \cite{jiang2023diabetes, peng2024gaxgboost}. Collectively, these studies demonstrate that boosting based methods consistently achieve strong performance on large population datasets such as NHANES. However, despite differences in algorithms and cohorts, they almost universally formulate diabetes prediction as a binary classification problem, leaving intermediate metabolic states comparatively underexplored.

\subsection{Machine Learning for Early Detection of Prediabetes}

Recent work has increasingly focused on identifying prediabetes before it progresses to established diabetes. Kopitar et al. used electronic health record data to detect impaired fasting glucose and undiagnosed diabetes \cite{kopitar2020early}, Kulkarni et al. used electrocardiography measurements to detect patterns associated with prediabetes and diabetes \cite{ecgdiabetes2023}, and Li et al. found that XGBoost could provide an interpretable estimate of prediabetes risk from population-survey variables \cite{li2025prediabetes}. More recent studies have treated prediabetes more directly as its own prediction target. Hu et al. developed a nomogram for predicting 5-year incident prediabetes in Chinese adults \cite{hu2023nomogram}, and Almadhoun and Burhanuddin compared several classifiers for distinguishing normal from prediabetic cases \cite{almadhoun2025prediabetes}. Fan et al. extended this direction to multiclass diagnosis by separating non-diabetic, prediabetic, and diabetic patients with an interpretable residual-network model \cite{fan2026progmdd}. Their analysis also showed that prediabetes overlaps with neighboring glycemic states in the raw feature space, although their model, which used HbA1c as an input, still separated the three classes with high accuracy. However, many approaches either predict future onset, combine abnormal glycemic states into a single group, or retain highly diagnostic biomarkers such as HbA1c as input features.

\subsection{Transformer-Based and Attention-Based Models for Diabetes Prediction}

Recent studies have explored transformer-based architectures for diabetes prediction using structured clinical data. TabTransformer has been applied to binary diabetes classification using tabular medical variables \cite{mukherjee2026xaiiot, kevin2025diabetes}, and tabular transformers have been extended to more specialized settings, including longitudinal clinical and DXA-derived features \cite{khan2025early} and metagenomic profiles \cite{wang2025msft}. Oba et al. applied TabNet to annual health-screening records to predict whether Type 2 diabetes status would worsen over time \cite{oba2021interpretable}. Their results also showed that strong boosting methods remained competitive, suggesting that attention-based models do not necessarily outperform conventional ensemble methods on structured healthcare data. It remains unclear whether transformer-based models can better capture the overlapping feature patterns associated with prediabetes while avoiding direct label leakage from HbA1c.

\subsection{Fairness and Bias in Healthcare Machine Learning}

Concerns about fairness are particularly relevant in the case of diabetes risk prediction. When certain populations are underrepresented in training datasets, prediction models may produce less reliable results for those groups, and several researchers have emphasized that evaluating algorithmic fairness should be considered an important step when developing machine learning systems for healthcare applications \cite{chinta2025fairness}. Existing prediction tools such as the Framingham Offspring Risk Score and the Prediabetes Risk Test have been shown to produce different levels of risk estimation for different populations when evaluated using large national datasets such as NHANES \cite{racialbiasdiabetes}. Other work has also highlighted the importance of examining error rates and calibration across demographic groups when evaluating healthcare prediction models \cite{chakradeo2025bias}. Fairness evaluation also involves inherent trade-offs, since different group-level criteria cannot always be satisfied simultaneously \cite{kleinberg2017tradeoffs}.

\section{Methods}
 
\subsection{Dataset}
 
We used data from the National Health and Nutrition Examination Survey (NHANES), a program conducted by the National Center for Health Statistics that combines interviews, physical examinations, and laboratory testing to assess the health and nutritional status of people in the United States \cite{nhanes}. For this study, we combined six NHANES cycles spanning 2007-2008 through 2017-2018. Data from the different NHANES components were merged using the participant identifier SEQN.
 
The dataset included demographic details such as age, sex, and race or ethnicity, body measurements like body mass index and waist circumference, blood pressure readings, and laboratory measurements including HbA1c, fasting glucose, and lipid-related variables. We also used questionnaire data that capture lifestyle factors such as smoking, physical activity, sedentary behavior, and family history of diabetes. Race and ethnicity data were taken from the RIDRETH1 variable and harmonized across cycles. This variable classifies participants as Mexican American, Other Hispanic, Non-Hispanic White, Non-Hispanic Black, or Other Race Including Multi-Racial, with Non-Hispanic Asian participants grouped within the Other Race category. For blood pressure, repeated systolic and diastolic measurements were averaged to get a single value for each participant. After merging all the selected components, the dataset contained a total of 19,307 participants.
 
Participants missing values in key variables, including age, body mass index, systolic blood pressure, diastolic blood pressure, and HbA1c, were removed. We also limited the dataset to adults aged 18 years and older. After applying these filtering steps, the final cohort used for the multiclass formulation included 15,356 individuals.
 
Glycemic status was defined using HbA1c thresholds based on standard clinical guidelines \cite{ada2026}. For the primary multiclass formulation, individuals with HbA1c below 5.7\% were labeled as normal, individuals with 5.7\% <= HbA1c < 6.5\% were labeled as prediabetic, and individuals with HbA1c >= 6.5\% were labeled as diabetic. This resulted in 9,249 normal cases, 4,279 prediabetes cases, and 1,828 diabetes cases, representing 60.2\%, 27.9\%, and 11.9\% of the final cohort, respectively.
 
Since HbA1c was used to define the outcome labels, it was excluded from the predictor set during model training to avoid label leakage. Fasting glucose was kept as a predictor, even though it is also a glycemic biomarker. Unlike HbA1c, it was not used to set the class boundaries, which means keeping it does not put the outcome definition back into the input features. HbA1c tracks average glucose over several months, whereas fasting glucose is a single reading taken at one point in time. Therefore it is connected to the labeling criterion without being equivalent to it, and it is the kind of measurement that is usually on hand in a screening setting. One complication is that NHANES measures fasting glucose only in the morning fasting subsample, so it is missing for a substantial proportion of participants. Rather than restricting the cohort to fasting participants, which would have shrunk the sample and altered its composition, we imputed missing values using the same median approach applied to the other continuous variables. A summary of the continuous and categorical variables used in this study is shown in Table \ref{tab:dataset_variables}, and the distribution of the target classes in the final dataset is presented in Table \ref{tab:class_distribution}.

\begin{table}[t]
\centering
\small
\begin{tabularx}{\columnwidth}{@{}p{2.2cm} p{2.2cm} >{\raggedright\arraybackslash}X@{}}
\hline
\textbf{Variable Type} & \textbf{Category} & \textbf{Variables} \\
\hline
 
Continuous Variables & Demographics & Age \\
& Body Measurements & Body Mass Index, Waist Circumference \\
& Blood Pressure & Systolic Blood Pressure, Diastolic Blood Pressure \\
& Laboratory & Fasting Glucose, Total Cholesterol, HDL, LDL, Triglycerides \\
& Lifestyle & Sedentary Behavior \\
 
\hline
 
Categorical Variables & Demographics & Sex, Race or Ethnicity \\
& Lifestyle & Smoking Status, Physical Activity Indicators \\
& Family History & Family History of Diabetes \\
 
\hline
\end{tabularx}
\caption{Continuous and categorical variables used in the dataset}
\label{tab:dataset_variables}
\end{table}

\begin{table}[t]
\centering
\begin{tabular}{lcc}
\hline
\textbf{Class} & \textbf{Samples} & \textbf{Percentage} \\
\hline
Normal & 9,249 & 60.2\% \\
Prediabetes & 4,279 & 27.9\% \\
Diabetes & 1,828 & 11.9\% \\
\hline
\end{tabular}
\caption{Class distribution}
\label{tab:class_distribution}
\end{table}

\subsection{Experimental Pipeline}
 
\subsubsection{Feature Processing and Encoding}
 
Missing values in continuous variables were filled using the median value of the corresponding feature, since median imputation is relatively robust to outliers and better preserves the distribution of skewed clinical variables compared to mean imputation. The median values were computed on the full cohort before the split between the train and test and since they do not use outcome information, this does not introduce label leakage. Although alternative imputation strategies may influence model behavior to some extent, evaluating sensitivity to different imputation methods was outside the scope of the current study. For categorical variables, missing entries were treated as a separate category rather than removing those samples from the dataset.
 
Variables such as sex, race or ethnicity, smoking behavior, physical activity indicators, and family history of diabetes were converted into numerical form using integer-based label encoding. Label encoding was selected primarily because several of the models used in this study are tree-based methods, which accommodate integer-encoded categorical variables directly without relying on Euclidean distances, and it also helped keep the feature space compact compared to one-hot encoding.
 
Continuous variables were standardized after preprocessing and feature construction. While tree-based models are generally less affected by feature scaling, standardization was retained because transformer based architectures benefited from normalized continuous features, while applying the same preprocessing to all models ensured consistency across experiments.
 
\subsubsection{Feature Engineering}
 
Additional features were created to capture metabolic and cardiovascular patterns that may not be fully reflected by the original variables alone. The triglyceride-to-HDL ratio was included because high triglycerides combined with low HDL cholesterol are often linked to insulin resistance. Although both variables were already included separately, the ratio captures their combined effect and may represent metabolic risk more directly. Pulse pressure was also calculated from systolic and diastolic blood pressure because the gap between these values may reflect vascular stress that is not fully captured by either measure alone. Interaction terms between age and body mass index, and between age and waist circumference, were added to capture how body composition may affect diabetes risk differently across age groups.
 
Several binary clinical indicators were also created, including obesity status, elevated blood pressure, elevated fasting glucose, high triglycerides, and low HDL cholesterol. Elevated fasting glucose ($\geq$100 mg/dL) and elevated blood pressure ($\geq$130/80 mmHg) followed American Diabetes Association (ADA) criteria \cite{ada2026}, while obesity was defined as BMI $\geq$30 kg/m$^2$, high triglycerides as $\geq$150 mg/dL, and low HDL cholesterol as $<$40 mg/dL. Although derived from existing variables, these indicators are not fully redundant because they represent clinically meaningful cut points that may help separate risk groups. The same engineered feature set was used across all prediction tasks to keep the multiclass and binary formulations comparable.
 
\subsubsection{Train-Test Splitting and Class Weighting}
 
The dataset was divided into training and testing sets using a 75/25 split, with stratified sampling applied to preserve the original class distribution across the glycemic categories. A fixed random seed of 42 was used throughout the experiments to support reproducibility.
 
Class imbalance was addressed using weights calculated from the training data, giving greater influence to underrepresented classes during model optimization, as recommended for learning from skewed class distributions \cite{he2009imbalanced}. Class weights were computed using the balanced formulation implemented in scikit-learn, where each class weight is inversely proportional to its frequency in the training data. Oversampling methods such as SMOTE were not applied because the goal was to evaluate model performance under the natural distribution of the NHANES cohort, and synthetic samples could introduce artificial combinations of demographic, metabolic, and lifestyle characteristics that may not reflect real participants.
 
\subsubsection{Experimental Formulations}
 
All prediction tasks in this study were built using the same preprocessing pipeline, feature set, train-test splitting strategy, and evaluation procedure. The main formulation treated glycemic prediction as a three-class classification problem involving normal, prediabetes, and diabetes categories.
 
To further examine the difficulty of identifying intermediate metabolic states, two additional binary settings were evaluated. The first focused only on distinguishing normal individuals from those with prediabetes, excluding diabetic participants from the dataset. It is intended as a controlled contrast that isolates the boundary between normal and prediabetic states, not as a standalone screening model, since diabetic individuals cannot be identified and excluded in advance in a real screening setting. Secondly, we combined prediabetes and diabetes into a single abnormal glycemic category and compared it against normal individuals in a broader screening-oriented setting.

\subsection{Modeling Approach}
 
\subsubsection{XGBoost}
 
We used XGBoost \cite{xgboost} as the primary model in this study because of its strong performance on structured tabular data and its ability to capture nonlinear relationships and feature interactions across demographic, clinical, and lifestyle variables.
 
Hyperparameters were selected by randomized search with 5-fold cross-validation, using a fixed seed of 42 and balanced sample weights during fitting. Table \ref{tab:hyperparam_search} reports the full search space. The binary formulations used shallower trees, lower learning rates, and a wider regularization space than the multiclass setting, and shared identical grids with each other. For the multiclass XGBoost model, the search was run for 40 iterations and optimized using macro F1. For the two binary formulations, the search was run for 80 iterations and optimized using positive-class F1, where the positive class was prediabetes in the normal versus prediabetes task and the combined abnormal group in the normal versus abnormal glycemic state task. Model tuning was done only on the training data, and final performance was evaluated on the separate test set.
 
\begin{table}
\centering
\footnotesize
\resizebox{\columnwidth}{!}{
\begin{tabular}{lll}
\hline
\textbf{Hyperparameter} & \textbf{Multiclass} & \textbf{Binary Formulations} \\
\hline
Tree depth & 3, 4, 5 & 2, 3, 4 \\
Learning rate & 0.03, 0.05, 0.07 & 0.01, 0.03, 0.05 \\
Number of trees & 400, 600, 800 & 500, 700, 900 \\
Row subsampling & 0.8, 0.9, 1.0 & 0.7, 0.8, 0.9 \\
Column subsampling & 0.8, 0.9, 1.0 & 0.7, 0.8, 0.9 \\
Minimum child weight & 1, 3, 5 & 3, 5, 7, 10 \\
Minimum split gain & 0, 0.1, 0.3 & 0.1, 0.3, 0.5, 1.0 \\
L1 regularization & -- & 0, 0.5, 1, 2 \\
L2 regularization & -- & 1, 3, 5, 10 \\
Maximum delta step & -- & 0, 1, 3, 5 \\
\hline
\end{tabular}
}
\caption{Hyperparameter search space for the primary XGBoost model. Dashes indicate parameters not included in the corresponding search.}
\label{tab:hyperparam_search}
\end{table}

\subsubsection{Fairness-Aware Reweighting and Baseline Model}
 
We evaluated a fairness-aware weighting strategy for the primary multiclass XGBoost model, in which sample weights were adjusted using both class labels and race or ethnicity group membership. For comparison, a baseline multiclass XGBoost model was trained using the same data split and hyperparameter settings but with only standard class-based weighting. This allowed the impact of fairness-aware reweighting to be evaluated independently from other modeling changes.
 
\subsubsection{Comparison Models and Ensemble Methods}
 
LightGBM \cite{lightgbm} and CatBoost \cite{catboost} were included as additional gradient boosting models. To keep the comparison across boosting frameworks fair, both were tuned using the same randomized search procedure applied to XGBoost, with the XGBoost ranges mapped onto the corresponding hyperparameters in each library and the same cross-validation scheme, scoring metric, and class-weighting strategy. To limit computational cost, LightGBM and CatBoost used reduced grid density and sampling budgets. LightGBM sampled 25 configurations in the multiclass setting and 40 in the binary formulations, and CatBoost sampled 20 in all settings. Since LightGBM grows trees leaf-wise, the number of leaves was tuned jointly with tree depth. CatBoost has no counterpart to minimum split gain, maximum delta step, or L1 regularization, and its leaf-size constraint was omitted to limit search complexity. CatBoost also handled categorical variables directly through its native categorical feature processing. HistGradientBoosting, Random Forest \cite{breiman2001}, and Logistic Regression were not tuned and instead served as fixed baselines, offering reference points that do not depend on the tuning budget. Class imbalance was handled the same way across all models.
 
We also explored a soft voting ensemble approach by averaging the predicted class probabilities from the primary boosting models, with final predictions assigned based on the highest combined probability score.

\subsubsection{Transformer-Based Models for Tabular Data}
 
To broaden the comparison beyond conventional machine learning models, we evaluated three transformer-based architectures designed for structured tabular data, to examine whether attention mechanisms could capture relationships among demographic, clinical, and lifestyle variables that may not be fully represented by tree-based methods. TabTransformer uses self-attention to learn contextual representations of categorical variables \cite{huang2020tabtransformer}. FT-Transformer extends this idea by representing both categorical and continuous variables as feature tokens and applying attention across the full feature sequence \cite{gorishniy2021fttransformer}. SAINT further incorporates attention across both features and samples, allowing each patient representation to draw context from other records within the same batch \cite{somepalli2021saint}. We used a supervised SAINT variant without contrastive pre-training. Each transformer architecture was evaluated across all three prediction settings.
 
All transformer models used the same leakage-aware predictor set as the conventional models. Categorical variables were integer-encoded and mapped to learned embeddings, while continuous variables were standardized using statistics calculated from the training data. Class imbalance was addressed using weighted cross-entropy loss based on the class distribution in the training set. We optimized all transformer models with AdamW using a learning rate of $10^{-3}$, weight decay of $10^{-4}$, a batch size of 256, and a dropout rate of 0.2. Each model used a multilayer perceptron classification head, applied to the flattened categorical embeddings combined with the continuous variables for TabTransformer and to a learnable classification token for FT-Transformer and SAINT. The main architectural and training settings for each model are summarized in Table \ref{tab:transformer_architectures}.
 
\begin{table*}
\centering
\footnotesize
\resizebox{\textwidth}{!}{
\begin{tabular}{llll}
\hline
\textbf{Model} & \textbf{Feature Representation} & \textbf{Transformer Configuration} & \textbf{Training} \\
\hline
TabTransformer & 16-d categorical embeddings with continuous features & 2 encoder layers, 4 heads & 40 epochs, batch size 256 \\
FT-Transformer & 32-d categorical and continuous feature tokens & 3 encoder layers, 4 heads & 50 epochs, batch size 256 \\
SAINT & 32-d feature tokens with feature and sample attention & 3 attention stages, 4 heads & 50 epochs, batch size 256 \\
\hline
\end{tabular}
}
\caption{Summary of transformer model architectures and training configurations.}
\label{tab:transformer_architectures}
\end{table*}
 
\subsubsection{Model Interpretation}
 
To better understand model behavior, we used SHAP \cite{shap} to examine feature contributions, focusing primarily on the prediabetes class. SHAP values were computed on a subset of the test data to keep computation manageable while still capturing overall model behavior, and summary plots and mean absolute SHAP values were used to identify the most influential predictors. Because post-hoc explanations do not provide a complete or causal account of model behavior, and interpretability itself can be defined in different ways, the results were interpreted cautiously \cite{doshivelez2017rigorous, lipton2018mythos}. In particular, correlated metabolic and anthropometric features may redistribute importance across related variables, which can influence the ranking of individual predictors without necessarily changing the overall predictive behavior of the model.

\subsection{Evaluation Metrics}
 
Model performance was evaluated using accuracy, precision, recall, specificity, and F1 score. These measures are computed from the number of true positives (TP), true negatives (TN), false positives (FP), and false negatives (FN):
 
\begin{equation}
\text{Precision} = \frac{TP}{TP + FP}
\end{equation}
 
\begin{equation}
\text{Recall} = \frac{TP}{TP + FN}
\end{equation}
 
\begin{equation}
\text{Specificity} = \frac{TN}{TN + FP}
\end{equation}

Recall is equivalent to sensitivity. For the multiclass task, sensitivity and specificity are computed separately for each class, treating that class as positive and the remaining two as negative. The F1 score combines precision and recall as their harmonic mean:
 
\begin{equation}
\text{F1} = 2 \cdot \frac{\text{Precision} \cdot \text{Recall}}{\text{Precision} + \text{Recall}}
\end{equation}
 
Because the dataset is imbalanced, particularly for the diabetes category, we treat macro-averaged F1 as the primary metric:
 
\begin{equation}
\text{Macro-F1} = \frac{1}{N} \sum_{i=1}^{N} \text{F1}_i
\end{equation}

Here, $N$ is the number of classes and $\text{F1}_i$ is the F1 score of class $i$. Unlike weighted F1, macro F1 averages the per-class scores without weighting by class frequency, so every glycemic state carries equal weight and strong results on the majority normal class cannot mask weak detection of the smaller prediabetes and diabetes groups. We also report balanced accuracy, computed as the average recall across classes, as a complement to macro F1.
 
Class-wise precision, recall, and F1 scores were additionally reported, and confusion matrices were used to identify common misclassification patterns, particularly whether prediabetes cases were being confused with normal or diabetic samples. For the binary formulations, precision, recall, and F1 score were reported with respect to the positive class.
 
To examine the consistency of model performance across different data splits, we also performed 5-fold stratified cross-validation using macro F1 score as the primary metric, and approximate 95\% confidence intervals were calculated from the fold-level macro F1 scores. As shown in Table \ref{tab:multiclass_cv_stability}, the tree-based and boosting models performed similarly, with narrow and largely overlapping confidence intervals that point to stable behavior across different partitions of the dataset. Logistic Regression was the exception. Its mean macro F1 of 0.6662 sits below the other models and its confidence interval does not overlap with them, which fits the expectation that a linear model captures less of the nonlinear structure separating the three glycemic classes. Since the intervals overlap so heavily among the tree-based and boosting models, the small gaps in macro F1 among them are best read as descriptive rather than as firm evidence that any one model outperforms the rest.
 
\begin{table}[t]
\centering
\footnotesize
\begin{tabular}{lcc}
\hline
\textbf{Model} & \textbf{Macro F1} & \textbf{95\% Confidence Interval} \\
\hline
XGBoost & 0.6980 $\pm$ 0.0076 & [0.6913, 0.7047] \\
LightGBM & 0.6930 $\pm$ 0.0064 & [0.6874, 0.6986] \\
CatBoost & 0.6998 $\pm$ 0.0096 & [0.6914, 0.7082] \\
HistGradientBoosting & 0.6938 $\pm$ 0.0057 & [0.6888, 0.6988] \\
Random Forest & 0.6980 $\pm$ 0.0064 & [0.6924, 0.7036] \\
Logistic Regression & 0.6662 $\pm$ 0.0078 & [0.6594, 0.6730] \\
\hline
\end{tabular}
\caption{Cross-validation stability analysis using macro F1 score. Approximate 95\% confidence intervals were calculated from the 5-fold cross-validation results.}
\label{tab:multiclass_cv_stability}
\end{table}

\section{Results}
 
\subsection{Multiclass Model Performance}
 
We evaluated all models on a separate test set. Overall, performance was consistent across models, with accuracy ranging from 0.69 to 0.72 and macro F1 scores from 0.68 to 0.70. Balanced accuracy also remained close to standard accuracy across models, ranging from 0.71 to 0.72. This suggests that the reported performance was not driven only by the majority normal class, but reflected relatively balanced recall across the three glycemic groups.
 
\begin{figure*}[t]
\centering
 
\includegraphics[width=0.32\textwidth]{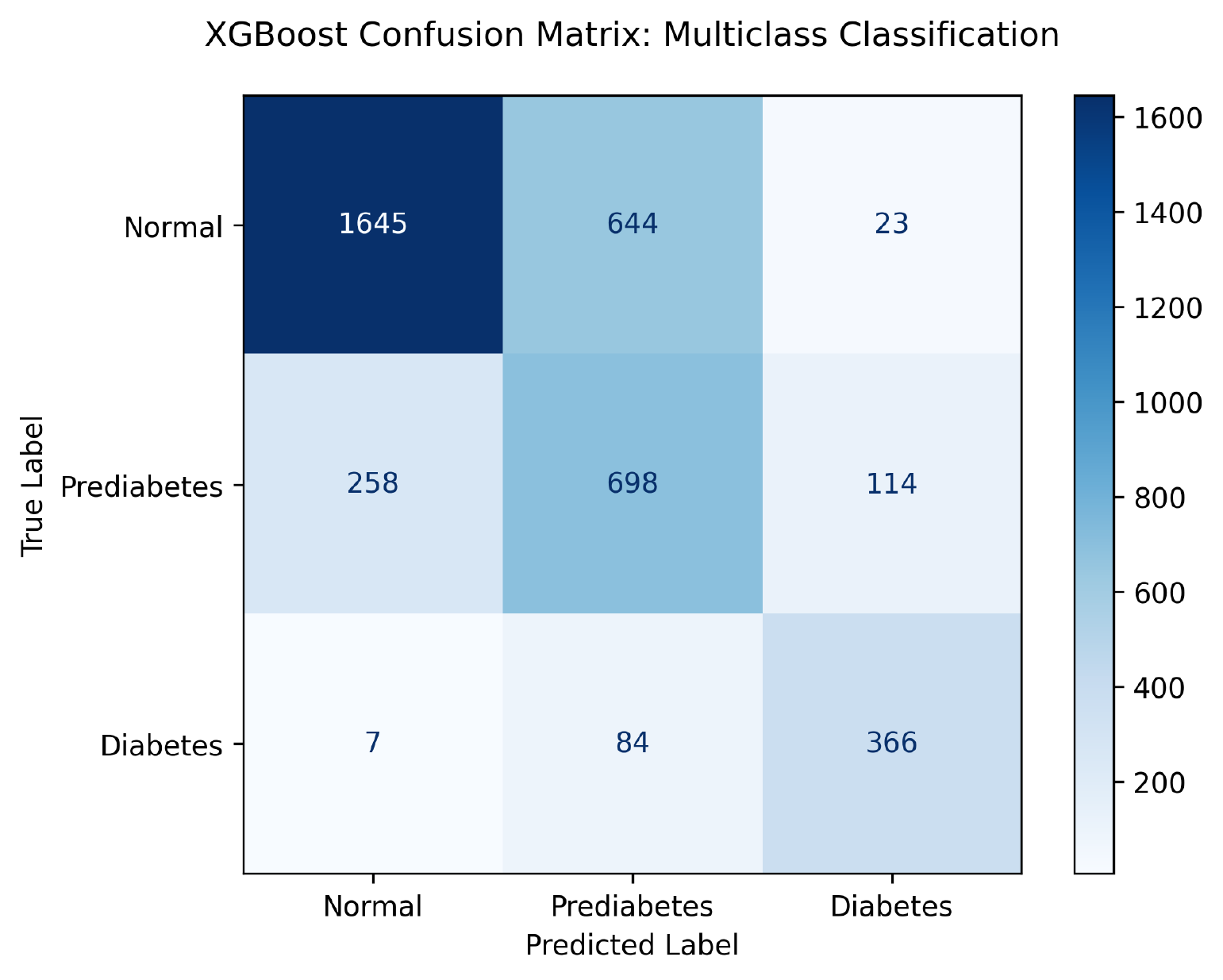}
\hfill
\includegraphics[width=0.32\textwidth]{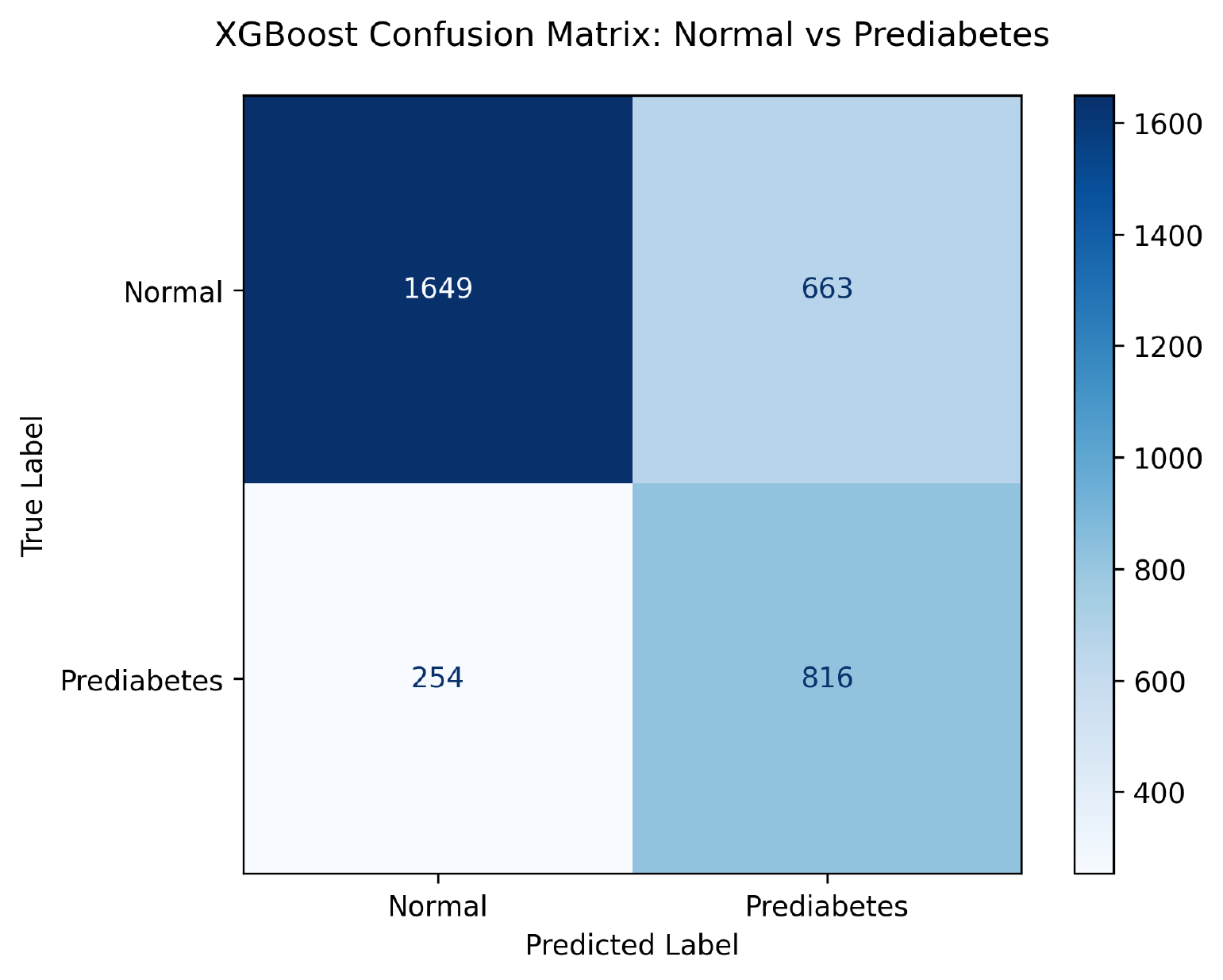}
\hfill
\includegraphics[width=0.32\textwidth]{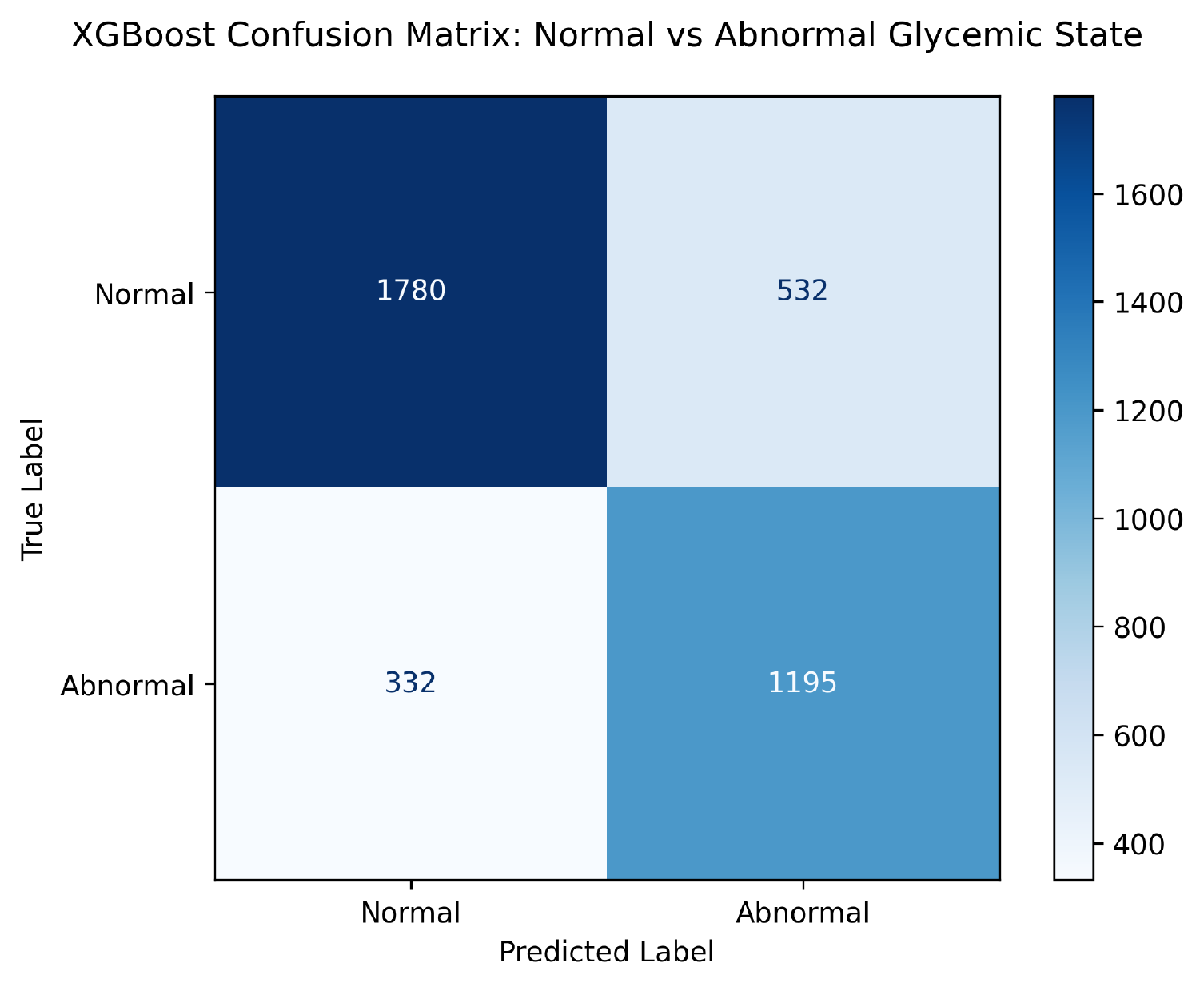}
 
\caption{Confusion matrices for XGBoost across the multiclass and binary prediction formulations.}
\label{fig:xgb_cm_comparison}
\end{figure*}

Full results for the boosting, ensemble, and baseline models are reported in Table \ref{tab:model_comparison}. XGBoost achieved an accuracy of 0.71, balanced accuracy of 0.72, and a macro F1 score of 0.70. The model performed well on the normal and diabetes classes, with strong precision and relatively balanced recall. As expected, prediabetes remained the most difficult class to classify. The model achieved a precision of 0.49, a sensitivity of 0.65 and a specificity of 0.74 for this class, reflecting the difficulty in separating prediabetes from both normal and diabetic cases without HbA1c. Although precision for the prediabetes class remained moderate, the relatively higher recall is important in a screening context, where failing to identify at-risk individuals may be more problematic than generating additional false positives. Most prediabetes misclassifications involved the normal class rather than the diabetes class, as shown in the confusion matrices in Figure \ref{fig:xgb_cm_comparison}. For diabetes, XGBoost reached a specificity of 0.96, so few individuals without diabetes were labeled diabetic.
 
The remaining models showed similar performance patterns. Once LightGBM and CatBoost were tuned under the same search procedure, both landed very close to XGBoost, each reaching an accuracy of 0.70 and a macro F1 of 0.69. For prediabetes, XGBoost, the soft voting ensemble, HistGradientBoosting, and Random Forest all reached an F1 of 0.56, with LightGBM and CatBoost just behind at 0.55. Sensitivity for this class varied a little more, from 0.62 for Random Forest up to 0.66 for HistGradientBoosting. The diabetes class was handled consistently, with sensitivity of 0.79 to 0.80 for XGBoost, LightGBM, CatBoost, and the ensemble, and specificity between 0.95 and 0.97 across all models. For the normal class, F1 held at 0.78 for the boosting models and the ensemble and rose slightly to 0.79 for Random Forest. No single boosting model was clearly best across every class, and the gaps between them were small.
 
\begin{table*}[!t]
\centering
\footnotesize
\resizebox{\textwidth}{!}{
\begin{tabular}{lccccccccccccccc}
\hline
\textbf{Model} & \textbf{Accuracy} & \textbf{Balanced Accuracy} & \textbf{Macro F1} & \textbf{Normal Precision} & \textbf{Normal Recall} & \textbf{Normal Specificity} & \textbf{Normal F1} & \textbf{Prediabetes Precision} & \textbf{Prediabetes Recall} & \textbf{Prediabetes Specificity} & \textbf{Prediabetes F1} & \textbf{Diabetes Precision} & \textbf{Diabetes Recall} & \textbf{Diabetes Specificity} & \textbf{Diabetes F1} \\
\hline
XGBoost & 0.71 & 0.72 & 0.70 & 0.86 & 0.71 & 0.83 & 0.78 & 0.49 & 0.65 & 0.74 & 0.56 & 0.73 & 0.80 & 0.96 & 0.76 \\
LightGBM & 0.70 & 0.72 & 0.69 & 0.86 & 0.71 & 0.82 & 0.78 & 0.48 & 0.64 & 0.74 & 0.55 & 0.71 & 0.79 & 0.96 & 0.75 \\
CatBoost & 0.70 & 0.71 & 0.69 & 0.86 & 0.71 & 0.83 & 0.78 & 0.48 & 0.64 & 0.73 & 0.55 & 0.70 & 0.79 & 0.95 & 0.74 \\
Ensemble & 0.70 & 0.72 & 0.70 & 0.86 & 0.71 & 0.83 & 0.78 & 0.49 & 0.65 & 0.73 & 0.56 & 0.72 & 0.80 & 0.96 & 0.75 \\
HistGradientBoosting & 0.70 & 0.71 & 0.70 & 0.86 & 0.71 & 0.82 & 0.78 & 0.49 & 0.66 & 0.73 & 0.56 & 0.73 & 0.77 & 0.96 & 0.75 \\
Random Forest & 0.72 & 0.71 & 0.70 & 0.84 & 0.75 & 0.79 & 0.79 & 0.50 & 0.62 & 0.76 & 0.56 & 0.76 & 0.77 & 0.97 & 0.76 \\
Logistic Regression & 0.69 & 0.71 & 0.68 & 0.85 & 0.70 & 0.82 & 0.77 & 0.47 & 0.64 & 0.73 & 0.54 & 0.69 & 0.79 & 0.95 & 0.73 \\
\hline
\end{tabular}
}
\caption{Performance comparison across the boosting, ensemble, and baseline models, including overall metrics, balanced accuracy, and class-wise precision, recall, specificity, and F1 scores for normal, prediabetes, and diabetes. Recall and specificity are computed for each class by treating that class as positive and the other two classes as negative.}
\label{tab:model_comparison}
\end{table*}

\subsection{Binary Formulations for Prediabetes Detection}
 
The two binary formulations produced different levels of difficulty. Relative to the multiclass setting, where prediabetes F1 ranged from approximately 0.54 to 0.56, the normal versus prediabetes formulation led to a modest improvement, with F1 scores ranging from 0.62 to 0.64. A larger increase was observed in the broader screening-oriented formulation, where prediabetes and diabetes were grouped into a single abnormal glycemic category. In this setting, F1 scores reached approximately 0.73. Simplifying the prediction task consistently improved performance across all evaluated models, and the substantially larger gain from combining prediabetes and diabetes indicates that the models more readily distinguish general metabolic abnormality than the finer distinction between normal and prediabetes. Table \ref{tab:binary_comparison} summarizes model performance across both formulations. Balanced accuracy followed the same pattern, ranging from 0.72 to 0.74 for normal versus prediabetes and from 0.77 to 0.78 for normal versus abnormal glycemic state. Specificity ranged from 0.70 to 0.77 in the normal versus prediabetes task and from 0.77 to 0.80 in the normal versus abnormal task, with Random Forest highest in both and XGBoost at 0.71 and 0.77, respectively.

\begin{table*}[!t]
\centering
\scriptsize
\resizebox{0.7\textwidth}{!}{
\begin{tabular}{llccccccc}
\hline
\textbf{Formulation} & \textbf{Model} & \textbf{Accuracy} & \textbf{Balanced Accuracy} & \textbf{Precision} & \textbf{Recall} & \textbf{Specificity} & \textbf{F1 Score} & \textbf{ROC-AUC} \\
\hline
Normal vs Prediabetes & XGBoost & 0.73 & 0.74 & 0.55 & 0.76 & 0.71 & 0.64 & 0.8182 \\
Normal vs Prediabetes & LightGBM & 0.73 & 0.74 & 0.55 & 0.76 & 0.71 & 0.64 & 0.8169 \\
Normal vs Prediabetes & CatBoost & 0.73 & 0.73 & 0.55 & 0.75 & 0.72 & 0.64 & 0.8144 \\
Normal vs Prediabetes & Ensemble & 0.73 & 0.74 & 0.55 & 0.76 & 0.72 & 0.64 & 0.8178 \\
Normal vs Prediabetes & HistGradientBoosting & 0.73 & 0.74 & 0.55 & 0.76 & 0.72 & 0.64 & 0.8169 \\
Normal vs Prediabetes & Random Forest & 0.74 & 0.72 & 0.58 & 0.68 & 0.77 & 0.62 & 0.8055 \\
Normal vs Prediabetes & Logistic Regression & 0.71 & 0.72 & 0.54 & 0.74 & 0.70 & 0.62 & 0.8016 \\
\hline
Normal vs Abnormal Glycemic State & XGBoost & 0.77 & 0.78 & 0.69 & 0.78 & 0.77 & 0.73 & 0.8674 \\
Normal vs Abnormal Glycemic State & LightGBM & 0.77 & 0.78 & 0.69 & 0.78 & 0.77 & 0.73 & 0.8690 \\
Normal vs Abnormal Glycemic State & CatBoost & 0.77 & 0.77 & 0.69 & 0.77 & 0.78 & 0.73 & 0.8698 \\
Normal vs Abnormal Glycemic State & Ensemble & 0.77 & 0.78 & 0.69 & 0.78 & 0.77 & 0.73 & 0.8696 \\
Normal vs Abnormal Glycemic State & HistGradientBoosting & 0.78 & 0.77 & 0.70 & 0.76 & 0.79 & 0.73 & 0.8672 \\
Normal vs Abnormal Glycemic State & Random Forest & 0.78 & 0.78 & 0.71 & 0.76 & 0.80 & 0.73 & 0.8622 \\
Normal vs Abnormal Glycemic State & Logistic Regression & 0.77 & 0.77 & 0.69 & 0.78 & 0.77 & 0.73 & 0.8574 \\
\hline
\end{tabular}
}
\caption{Performance comparison across targeted binary formulations. Metrics are reported for the positive class in each setup, corresponding to prediabetes in the first formulation and to the combined abnormal glycemic state, which includes both prediabetes and diabetes, in the second. Specificity is the proportion of normal individuals correctly classified as normal. Balanced accuracy is also reported to account for class imbalance between the normal and positive classes.}
\label{tab:binary_comparison}
\end{table*}

\subsection{Transformer-Based Results Across Prediction Formulations}
 
As shown in Table \ref{tab:transformer_formulations}, the transformer models produced results broadly comparable to the conventional machine learning baselines, but did not lead to a clear improvement in overall performance.

\begin{table*}[!t]
\centering
\footnotesize
\resizebox{\textwidth}{!}{
\begin{tabular}{llcccccccc}
\hline
\textbf{Formulation} & \textbf{Model} & \textbf{Accuracy} &
\textbf{Balanced Accuracy} & \textbf{Macro F1} &
\textbf{Precision} & \textbf{Recall} & \textbf{Specificity} &
\textbf{F1} & \textbf{ROC-AUC} \\
\hline
Multiclass & TabTransformer & 0.70 & 0.71 & 0.69 & 0.49 & 0.61 & 0.75 & 0.55 & 0.86 \\
Multiclass & FT-Transformer & 0.71 & 0.72 & 0.68 & 0.50 & 0.56 & 0.79 & 0.53 & 0.86 \\
Multiclass & SAINT & 0.68 & 0.71 & 0.67 & 0.46 & 0.61 & 0.72 & 0.53 & 0.85 \\
\hline
Normal vs Prediabetes & TabTransformer & 0.72 & 0.73 & 0.71 & 0.55 & 0.75 & 0.71 & 0.63 & 0.80 \\
Normal vs Prediabetes & FT-Transformer & 0.73 & 0.73 & 0.71 & 0.56 & 0.74 & 0.73 & 0.63 & 0.81 \\
Normal vs Prediabetes & SAINT & 0.72 & 0.73 & 0.70 & 0.54 & 0.77 & 0.70 & 0.63 & 0.81 \\
\hline
Normal vs Abnormal Glycemic State & TabTransformer & 0.76 & 0.77 & 0.76 & 0.67 & 0.79 & 0.74 & 0.73 & 0.86 \\
Normal vs Abnormal Glycemic State & FT-Transformer & 0.77 & 0.78 & 0.77 & 0.68 & 0.79 & 0.76 & 0.73 & 0.87 \\
Normal vs Abnormal Glycemic State & SAINT & 0.77 & 0.78 & 0.77 & 0.68 & 0.80 & 0.75 & 0.74 & 0.87 \\
\hline
\end{tabular}
}
\caption{Transformer-based model performance across the multiclass, normal versus prediabetes, and normal versus abnormal glycemic state formulations. For the multiclass task, precision, recall, specificity, and F1 refer to the prediabetes class, with specificity computed by treating prediabetes as positive and the remaining two classes as negative. For the binary tasks, these metrics refer to prediabetes in the normal versus prediabetes formulation and to the combined abnormal glycemic state in the normal versus abnormal formulation, and specificity is the proportion of normal individuals correctly classified as normal.}
\label{tab:transformer_formulations}
\end{table*}

Among the transformer models, TabTransformer achieved the highest macro F1 score of 0.69 and the highest prediabetes F1 score of 0.55. FT-Transformer achieved the highest overall accuracy and balanced accuracy, reaching 0.71 and 0.72, respectively. SAINT achieved a balanced accuracy of 0.71, although its overall accuracy and macro F1 score were slightly lower than those of the other transformer models.
 
Prediabetes remained the most difficult class across all three architectures, with F1 scores ranging from 0.53 to 0.55. Under the normal versus prediabetes formulation, prediabetes F1 increased to approximately 0.63 for all three transformer models. Here, FT-Transformer achieved the highest accuracy at 0.73, while SAINT achieved the highest prediabetes recall at 0.77. Performance improved further under the normal versus abnormal formulation. Abnormal-class F1 reached 0.73 for TabTransformer and FT-Transformer and 0.74 for SAINT. FT-Transformer and SAINT both reached the highest accuracy at 0.77 and the highest balanced accuracy at 0.78, and both attained a ROC-AUC of 0.87. Specificity ranged from 0.72 to 0.79 for prediabetes in the multiclass task, from 0.70 to 0.73 in the normal versus prediabetes task, and from 0.74 to 0.76 in the normal versus abnormal task. None of the three architectures consistently outperformed the boosting models, and all struggled to separate prediabetes from adjacent glycemic states, indicating that increasing model complexity alone does not overcome the intrinsic overlap between intermediate metabolic states in the available feature space.

\subsection{Effect of Problem Formulation}
 
Comparing the three formulations shows that the main difficulty in detecting prediabetes comes from overlap between neighboring glycemic states. As shown in Figure \ref{fig:xgb_cm_comparison}, diabetes is identified more cleanly than prediabetes in the multiclass setting, while most of the uncertainty appears around the normal and prediabetes boundary. This suggests that the model captures stronger metabolic signals for diabetes, but has more difficulty separating early-stage risk from normal variation.
 
The binary normal versus prediabetes formulation makes this issue clearer. By removing diabetes from the task, the model focuses only on early-stage detection, but the confusion matrix still shows overlap between the two groups. The model correctly identifies 816 of the 1,070 prediabetes cases, a sensitivity of 0.76, but it also classifies 663 of the 2,312 normal individuals as prediabetic, giving a specificity of 0.71. This pattern matches the results in Table \ref{tab:binary_comparison}, where sensitivity for prediabetes is stronger than precision, and shows that the boundary remains difficult to define.
 
The normal versus abnormal glycemic state formulation shows the clearest separation because prediabetes and diabetes are combined into a single positive class. This reduces the burden on the model by asking it to detect broader metabolic abnormality rather than distinguish between intermediate and advanced glycemic states. Here XGBoost reaches a sensitivity of 0.78 and a specificity of 0.77, so most abnormal cases are detected without misclassifying a large share of normal individuals. The progressive improvement observed when moving from multiclass prediction to broader abnormality screening suggests that the principal source of error lies in distinguishing intermediate metabolic states rather than detecting glycemic abnormality in general.
 
\subsection{UMAP Feature-Space Visualization}
 
\begin{figure}
\centering
\includegraphics[width=0.46\textwidth]{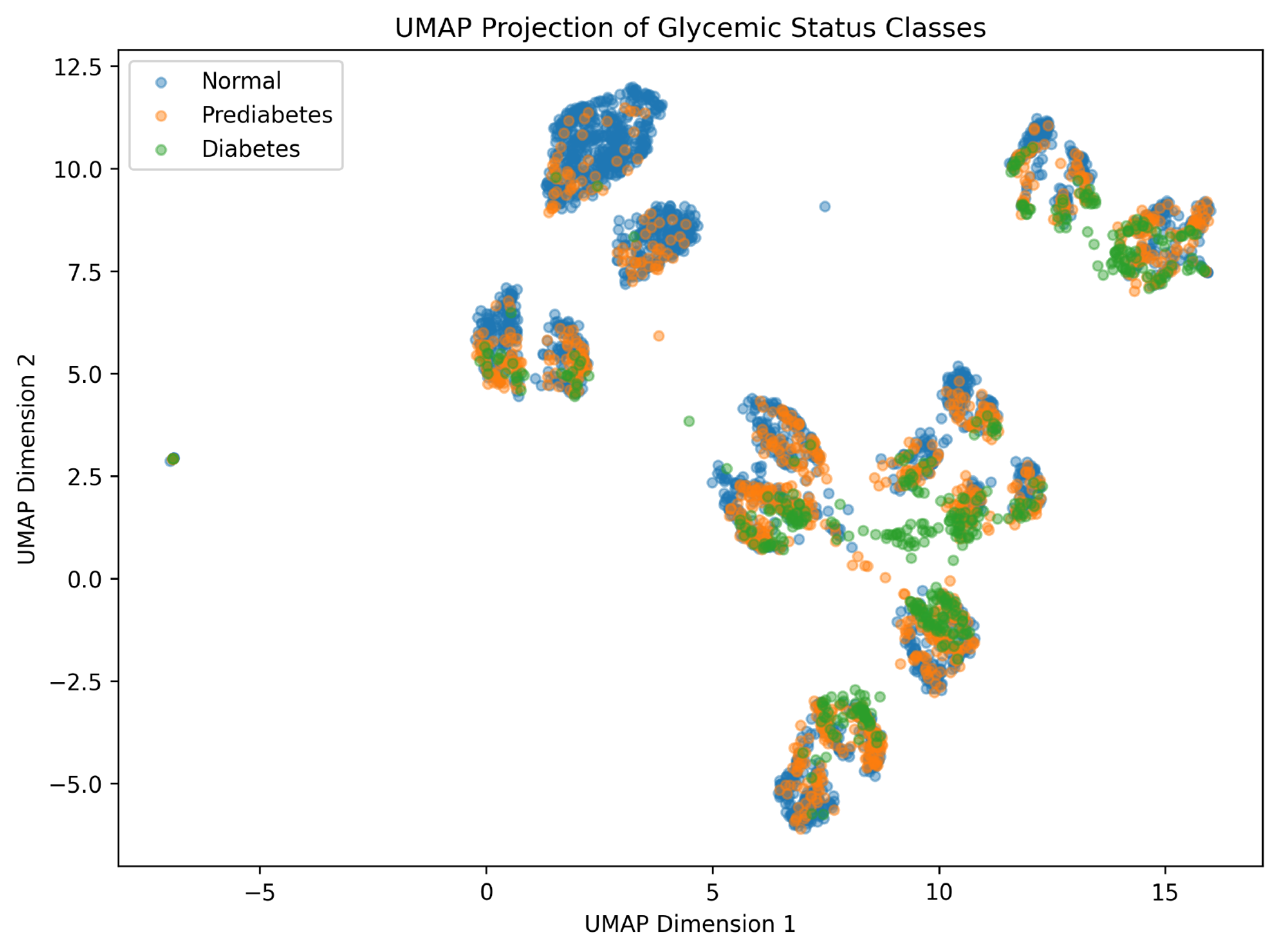}
\caption{UMAP visualization of the normal, prediabetes, and diabetes groups in the test data. HbA1c and participant identifiers were excluded, numeric variables were standardized, and categorical variables were one-hot encoded. UMAP was fitted on the training set and the held-out test set was projected into two dimensions, using 30 neighbors, a minimum distance of 0.1, Euclidean distance, and a random seed of 42.}
\label{fig:umap_projection}
\end{figure}

To examine how the glycemic classes were distributed in the feature space, we applied Uniform Manifold Approximation and Projection (UMAP) \cite{mcinnes2020umapuniformmanifoldapproximation} to the same leakage-aware predictor set used for classification, with settings given in the caption of Figure \ref{fig:umap_projection}.
 
As shown in Figure \ref{fig:umap_projection}, the projection contains several local groupings, but these groups do not clearly correspond to the three glycemic classes. Some regions contain higher concentrations of normal or diabetic participants, although substantial mixing remains throughout the projection. Prediabetes is particularly dispersed across regions containing both normal and diabetic participants rather than forming a separate group. This pattern lines up with the lower F1 scores that prediabetes produced across the models. We read this agreement as supporting evidence rather than a direct explanation, because UMAP is a nonlinear, exploratory projection that does not preserve the exact geometry of the feature space and can stretch or compress distances. For that reason, the groupings seen here should not be taken as distinct clinical subgroups.
 
\subsection{Encoding Sensitivity Analysis}
To examine whether the categorical encoding strategy affected the main multiclass results, we compared the primary XGBoost model using label encoding, one-hot encoding, and target encoding. As shown in Table \ref{tab:encoding_sensitivity}, performance changed only marginally across the three methods, so the main conclusions of our study hold regardless of how categorical variables are represented. The prediabetes F1 stayed modest under every encoding, which points back to feature overlap rather than encoding choice as the source of the difficulty.
 
\begin{table}[t]
\centering
\footnotesize
\resizebox{\columnwidth}{!}{
\begin{tabular}{lcccc}
\hline
\textbf{Encoding} & \textbf{Accuracy} & \textbf{Balanced Accuracy} & \textbf{Macro F1} & \textbf{Prediabetes F1} \\
\hline
Label & 0.71 & 0.72 & 0.70 & 0.56 \\
One-Hot & 0.71 & 0.72 & 0.69 & 0.55 \\
Target & 0.72 & 0.72 & 0.70 & 0.56 \\
\hline
\end{tabular}
}
\caption{Encoding sensitivity analysis for the multiclass XGBoost model.}
\label{tab:encoding_sensitivity}
\end{table}

\subsection{Fairness-Aware Reweighting}
 
\begin{figure}
\centering
\includegraphics[width=0.44\textwidth]{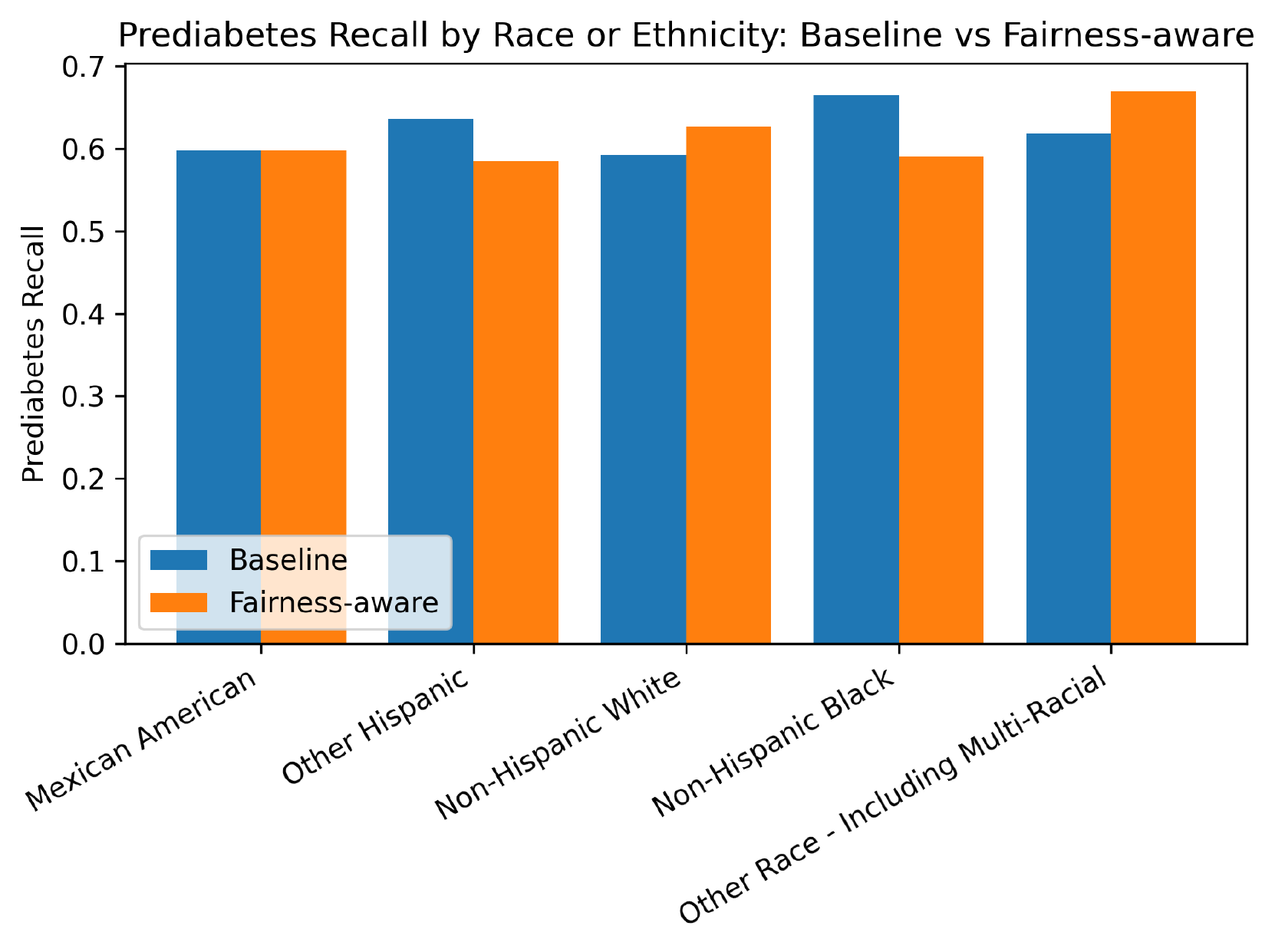}
\caption{Prediabetes recall across race or ethnicity groups for the baseline and fairness-aware multiclass XGBoost models.}
\label{fig_fairness}
\end{figure}

We also assessed how fairness-aware reweighting affected the consistency of prediabetes detection across racial groups in the multiclass setting. In this context, the recall of subgroups captures the proportion of true prediabetes cases within a given group that the model correctly identified. Relative to the baseline, the fairness-aware XGBoost model did not distribute prediabetes recall more evenly across race groups. Recall increased for Non-Hispanic White and Other Race participants but decreased for Other Hispanic and Non-Hispanic Black participants, while Mexican American participants stayed about the same, so the gap between the highest and lowest groups remained similar. The analysis also considered only recall rather than a fuller range of fairness measures. The results are therefore best read as exploratory evidence that reweighting alone shifts recall between groups rather than equalizing it, not as a comprehensive fairness assessment spanning all groups and error types. Figure \ref{fig_fairness} reports prediabetes recall across race groups for the baseline and fairness-aware models.
 
\section{Model Interpretation}
 
\begin{figure}
	\centering
	\includegraphics[width=0.32\textwidth]{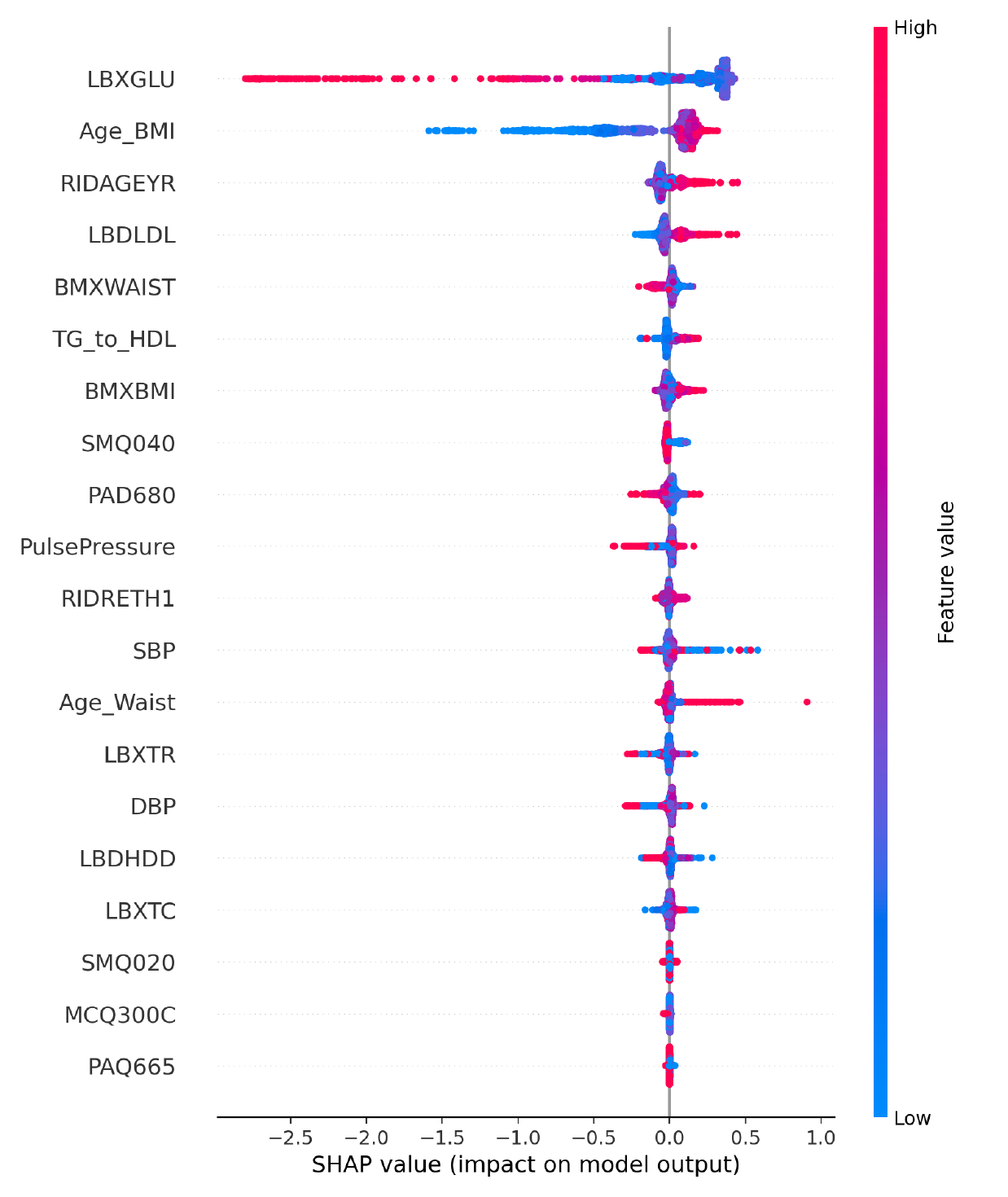}
	\caption{SHAP summary plot illustrating feature contributions to the prediabetes class in the XGBoost model. Laboratory features are LBXGLU for fasting glucose, LBDLDL for LDL cholesterol, LBDHDD for HDL cholesterol, LBXTR for triglycerides, and LBXTC for total cholesterol. Body and blood pressure features are BMXBMI for body mass index, BMXWAIST for waist circumference, and SBP and DBP for systolic and diastolic blood pressure. Demographic and lifestyle features are RIDAGEYR for age, RIDRETH1 for race or ethnicity, SMQ040 for current cigarette smoking, SMQ020 for having smoked at least 100 cigarettes in life, PAD680 for minutes of sedentary activity, PAQ665 for moderate recreational activities, and MCQ300C for a close relative having diabetes. Engineered features are Age\_BMI and Age\_Waist for the interactions of age with body mass index and waist circumference, TG\_to\_HDL for the triglyceride-to-HDL ratio, and PulsePressure for pulse pressure.}
	\label{fig_shap}%
\end{figure}

Figure \ref{fig_shap} shows the SHAP summary plot for the prediabetes class in the XGBoost model, where features are ordered by their mean absolute SHAP value and the spread of values reflects their influence on individual predictions. The most influential predictors include fasting glucose, the interaction between age and body mass index, age, and LDL cholesterol, followed by waist circumference, the triglyceride to HDL ratio, and body mass index. Fasting glucose is the strongest predictor by a wide margin. It is not used to define the labels, but it is closely related to HbA1c, so part of its contribution reflects glycemic information the two measures share.

The effect of fasting glucose is not uniform across its range. Very high values strongly push predictions away from prediabetes, which is consistent with these individuals being classified as diabetic instead, while low to moderate values contribute modestly toward prediabetes. The interaction between age and body mass index shows a clear pattern, which is that low values reduce predicted risk and high values increase it. Age and LDL cholesterol show consistent positive contributions, with higher values associated with increased predicted risk, which aligns with established clinical patterns.
 
Lipid variables also contribute to the predictions. A higher triglyceride-to-HDL ratio is associated with increased risk, while higher HDL cholesterol tends to reduce it. This reflects known mechanisms of metabolic dysfunction, where elevated triglycerides and reduced HDL are linked to insulin resistance. Waist circumference, pulse pressure, and systolic and diastolic blood pressure show smaller and less consistent effects. For waist circumference, higher values tend to reduce predicted risk, which most likely reflects information already captured by the age interaction terms rather than a protective effect. Physical activity and sedentary behavior variables contribute relatively little, and the binary clinical threshold flags do not appear among the 20 most influential features, which suggests that their information is already represented by the continuous measurements.
 
Overall, the SHAP analysis shows that the model relies mainly on glycemic, age, and body size measures, with lipid and lifestyle features playing a smaller role. The major predictors identified by SHAP exhibit continuous rather than discrete relationships with risk. This supports the view that glycemic status reflects a metabolic continuum rather than sharply separated categories.

\section{Discussion}
 
The findings show that both conventional machine learning models and transformer-based architectures perform well in predicting normal, prediabetes, and diabetes using NHANES data, with macro F1 scores around 0.70 across the boosting and ensemble models. This shows that the selected demographic, clinical, and lifestyle variables carry enough information to distinguish among the three glycemic states even without using HbA1c as an input feature.
 
None of the transformer-based models clearly outperformed the strongest conventional baselines, so attention-based architectures alone do not overcome the difficulty of separating the intermediate glycemic state. This agrees with Oba et al. \cite{oba2021interpretable}, where XGBoost and CatBoost outperformed the attention-based TabNet on health screening records. Given the close performance of conventional and transformer-based approaches, further gains likely depend more on richer feature sets or improved cohort design than on switching architectures.
 
Prediabetes remained the hardest class to predict across models, which reflects the biological continuum between normal glycemic regulation and diabetes, where fixed clinical thresholds are applied to a gradual process of metabolic change. Individuals near the normal-prediabetes or prediabetes-diabetes boundaries may therefore have similar demographic, metabolic, and lifestyle profiles despite being assigned to different classes. Fan et al. \cite{fan2026progmdd} reported a prediabetes F1 of 0.88, but with HbA1c among the inputs and only eight prediabetic test cases, suggesting that much of the difference reflects excluding HbA1c from the predictors rather than limitations in model capacity.
 
Fairness-aware reweighting did not make prediabetes recall more consistent across race groups, and instead shifted it between groups. In a healthcare context, this is important because uneven performance across demographic groups can translate into unequal opportunities for early detection and prevention. This suggests that reweighting alone is not enough, and that fairness considerations belong in model development itself, not only in evaluation performed after the fact.
 
Across the evaluated model families and prediction formulations, the overlap between adjacent glycemic states contributes more to prediction difficulty than differences in model architecture. The normal versus abnormal formulation reached an area under the curve of 0.87, compared with 0.82 for binary diabetes prediction from NHANES lifestyle variables \cite{riverosperez2025nhanes} and 0.90 in a high-risk NHANES subgroup \cite{yang2026nhanes_highrisk}, although targets and predictors differ across these studies. The models are therefore better at detecting general glycemic abnormality than at isolating prediabetes as a distinct intermediate state. One limitation of the current formulation is that standard multiclass classification treats normal, prediabetes, and diabetes as separate categories, even though they represent an ordered metabolic progression. Not all classification errors are equally serious, since misclassifying a prediabetic individual as normal is less severe than misclassifying a diabetic individual as normal, yet conventional multiclass approaches penalize both equally. Future work could explore ordinal classification or cost-sensitive learning to better reflect this structure.

\section{Limitations}
 
First, the outcome labels were defined using HbA1c thresholds, which is a practical and commonly used approach, but it does not fully reflect how diabetes and prediabetes are assessed in clinical practice, where diagnosis often involves multiple tests, medical history, and repeated measurements over time. In addition, NHANES is cross-sectional, so the models developed here focus on classification and do not provide insight into how individuals transition from prediabetes to diabetes, which may also obscure early predictive signals. Excluding HbA1c, while necessary to prevent leakage, may also remove a clinically dominant predictor, thereby lowering achievable performance. However, fasting glucose remained in the predictor set despite its biological relationship with HbA1c and may partially contribute to reconstructing glycemic status. It is also measured only in the fasting subsample, so imputed values were used elsewhere. Since it was the strongest predictor in the SHAP analysis, a check restricted to participants with measured fasting glucose would clarify how far the reported importance depends on imputation. External validation on an independent cohort was not performed, limiting the generalizability of the findings.
 
The fairness-aware reweighting analysis remained exploratory and was limited to the primary multiclass XGBoost model. It focused mainly on subgroup recall and did not evaluate other fairness-related outcomes such as subgroup precision, false positive rates, calibration, or intersectional differences across combined demographic categories. Finally, the study relies on the variables available in NHANES and evaluates a selected group of conventional and transformer-based models.

\section{Conclusion and Future Work}
 
This study examined how well conventional machine learning models and transformer-based architectures can classify individuals as normal, prediabetic, or diabetic using NHANES data. Across the boosting models XGBoost, LightGBM, and CatBoost, comparable performance was achieved within all evaluation metrics. Even without using HbA1c as an input feature, the models were able to distinguish between the three groups reasonably well. Prediabetes remained the most difficult class, although recall for this group was still relatively strong, which is important for early detection. Specificity for diabetes stayed between 0.95 and 0.97 across models, so few individuals without diabetes were misclassified as diabetic. 
 
Problem formulation mattered more than model choice, particularly when modeling intermediate disease states such as prediabetes. Among both model families, performance was strongest when prediabetes and diabetes were grouped together, meaning the models are better at recognizing general glycemic abnormality than identifying prediabetes as a separate intermediate state. Several extensions to this work are possible. Adding features such as dietary habits, medication use, or more detailed clinical variables may help improve separation between classes, especially for prediabetes. Longitudinal modeling and clinically actionable risk thresholds for early-stage detection are worth evaluating as well. Applying these models to external datasets, and ideally to longitudinal data, would show how well they generalize and whether they can support practical use in real-world settings.

\section{Data Availability}

The data used in this study are derived from the National Health and Nutrition Examination Survey, a public resource made available by the National Center for Health Statistics \cite{nhanes}. This study used the 2007-2008 to 2017-2018 cycles.

\section{Acknowledgements}
The authors thank the National Center for Health Statistics for making the NHANES data publicly available. This study is a secondary analysis of de-identified NHANES data. The Research Ethics Review Board of the National Center for Health Statistics approved NHANES and written informed consent was obtained from all adult participants. No further ethical approval was needed. This research did not receive specific grants from funding agencies in the public, commercial or non-profit sectors. The authors declare that they have no known competing financial interests or personal relationships that could have appeared to influence the work reported in this paper.




\bibliographystyle{elsarticle-num}
\bibliography{references}

@article{jiang2023diabetes,
  author  = {Jiang, Liangjun and Xia, Zhenhua and Zhu, Ronghui and Gong, Haimei and Wang, Jing and Li, Juan and Wang, Lei},
  title   = {Diabetes risk prediction model based on community follow-up data using machine learning},
  journal = {Preventive Medicine Reports},
  volume  = {35},
  pages   = {102358},
  year    = {2023},
  doi     = {10.1016/j.pmedr.2023.102358}
}

@article{peng2024gaxgboost,
  author  = {Li, Wenguang and Peng, Yan and Peng, Ke},
  title = {Diabetes prediction model based on {GA-XGBoost} and stacking ensemble algorithm},
  journal = {PLOS ONE},
  volume  = {19},
  number  = {9},
  pages   = {e0311222},
  year    = {2024},
  doi     = {10.1371/journal.pone.0311222}
}

@inproceedings{xu2019rf_xgboost,
  author    = {Xu, Zhongxian and Wang, Zhiliang},
  booktitle = {2019 Eleventh International Conference on Advanced Computational Intelligence (ICACI)},
  title     = {A Risk Prediction Model for Type 2 Diabetes Based on Weighted Feature Selection of Random Forest and {XGBoost} Ensemble Classifier},
  address   = {Guilin, China},
  year      = {2019},
  pages     = {278--283},
  doi       = {10.1109/ICACI.2019.8778622}
}

@article{healthcare_xgboost_2020,
  author  = {Deberneh, Henock M. and Kim, Intaek},
  title   = {Prediction of Type 2 Diabetes Based on Machine Learning Algorithm},
  journal = {International Journal of Environmental Research and Public Health},
  volume  = {18},
  number  = {6},
  pages   = {3317},
  year    = {2021},
  doi     = {10.3390/ijerph18063317}
}

@article{kopitar2020early,
  author  = {Kopitar, Leon and Kocbek, Primo{\v{z}} and Cilar, Leona and Sheikh, Aziz and {\v{S}}tiglic, Gregor},
  title   = {Early detection of type 2 diabetes mellitus using machine learning-based prediction models},
  journal = {Scientific Reports},
  volume  = {10},
  pages   = {11981},
  year    = {2020},
  doi     = {10.1038/s41598-020-68771-z}
}

@article{ecgdiabetes2023,
  author  = {Kulkarni, Anoop R. and Patel, Ashwini A. and Pipal, Kanchan V. and Jaiswal, Sujeet G. and Jaisinghani, Manisha T. and Thulkar, Vidya and Gajbhiye, Lumbini and Gondane, Preeti and Patel, Archana B. and Mamtani, Manju and Kulkarni, Hemant},
  title   = {Machine-learning algorithm to non-invasively detect diabetes and pre-diabetes from electrocardiogram},
  journal = {BMJ Innovations},
  volume  = {9},
  number  = {1},
  pages   = {32--42},
  year    = {2023},
  doi     = {10.1136/bmjinnov-2021-000759}
}

@article{li2025prediabetes,
  author  = {Li, Xiaolong and Ding, Fan and Zhang, Lu and Zhao, Shi and Hu, Zengyun and Ma, Zhanbing and Li, Feng and Zhang, Yuhong and Zhao, Yi and Zhao, Yu},
  title   = {Interpretable machine learning method to predict the risk of pre-diabetes using a national-wide cross-sectional data: evidence from {CHNS}},
  journal = {BMC Public Health},
  volume  = {25},
  number  = {1},
  pages   = {1145},
  year    = {2025},
  doi     = {10.1186/s12889-025-22419-7}
}

@article{chinta2025fairness,
  author  = {Chinta, Sribala Vidyadhari and Wang, Zichong and Palikhe, Avash and Zhang, Xingyu and Kashif, Ayesha and Smith, Monique Antoinette and Liu, Jun and Zhang, Wenbin},
  title   = {{AI}-driven healthcare: A review on ensuring fairness and mitigating bias},
  journal = {PLOS Digital Health},
  volume  = {4},
  number  = {5},
  pages   = {e0000864},
  year    = {2025},
  doi     = {10.1371/journal.pdig.0000864}
}

@article{racialbiasdiabetes,
  author  = {Cronj{\'e}, H{\'e}l{\'e}ne T. and Katsiferis, Alexandros and Elsenburg, Leonie K. and Andersen, Thea O. and Rod, Naja H. and Nguyen, Tri-Long and Varga, Tibor V.},
  title   = {Assessing racial bias in type 2 diabetes risk prediction algorithms},
  journal = {PLOS Global Public Health},
  volume  = {3},
  number  = {5},
  pages   = {e0001556},
  year    = {2023},
  doi     = {10.1371/journal.pgph.0001556}
}

@article{chakradeo2025bias,
  author  = {Chakradeo, Kaustubh and Huynh, Inchuen and Balaganeshan, Sedrah B. and Dollerup, Ole L. and Gade-J{\o}rgensen, Hjordis and Laupstad, Susanne K. and Malham, Mikkel and Nguyen, Tri-Long and Hulman, Adam and Varga, Tibor V.},
  title   = {Navigating fairness aspects of clinical prediction models},
  journal = {BMC Medicine},
  volume  = {23},
  number  = {1},
  pages   = {567},
  year    = {2025},
  doi     = {10.1186/s12916-025-04340-3}
}

@misc{nhanes,
  author       = {{National Center for Health Statistics}},
  title        = {National Health and Nutrition Examination Survey},
  howpublished = {\url{https://wwwn.cdc.gov/nchs/nhanes/}},
  note         = {Accessed 9 October 2026},
  year         = {2026}
}

@inproceedings{xgboost,
  author    = {Chen, Tianqi and Guestrin, Carlos},
  title     = {{XGBoost}: A Scalable Tree Boosting System},
  booktitle = {Proceedings of the 22nd ACM SIGKDD International Conference on Knowledge Discovery and Data Mining},
  publisher = {ACM},
  pages     = {785--794},
  year      = {2016},
  doi       = {10.1145/2939672.2939785}
}

@inproceedings{lightgbm,
  author    = {Ke, Guolin and Meng, Qi and Finley, Thomas and Wang, Taifeng and Chen, Wei and Ma, Weidong and Ye, Qiwei and Liu, Tie-Yan},
  title     = {{LightGBM}: A Highly Efficient Gradient Boosting Decision Tree},
  booktitle = {Proceedings of the 31st International Conference on Neural Information Processing Systems},
  publisher = {Curran Associates Inc.},
  address   = {Red Hook, NY, USA},
  pages     = {3149--3157},
  year      = {2017}
}

@inproceedings{catboost,
  author    = {Prokhorenkova, Liudmila and Gusev, Gleb and Vorobev, Aleksandr and Dorogush, Anna Veronika and Gulin, Andrey},
  title     = {{CatBoost}: Unbiased Boosting with Categorical Features},
  booktitle = {Proceedings of the 32nd International Conference on Neural Information Processing Systems},
  publisher = {Curran Associates Inc.},
  address   = {Red Hook, NY, USA},
  pages     = {6639--6649},
  year      = {2018}
}

@inproceedings{shap,
  author    = {Lundberg, Scott M. and Lee, Su-In},
  title     = {A Unified Approach to Interpreting Model Predictions},
  booktitle = {Proceedings of the 31st International Conference on Neural Information Processing Systems},
  publisher = {Curran Associates Inc.},
  address   = {Red Hook, NY, USA},
  pages     = {4768--4777},
  year      = {2017}
}

@article{breiman2001,
  author    = {Breiman, Leo},
  title     = {Random Forests},
  journal   = {Machine Learning},
  volume    = {45},
  number    = {1},
  pages     = {5--32},
  year      = {2001},
  publisher = {Springer},
  doi       = {10.1023/A:1010933404324}
}

@article{riverosperez2025nhanes,
  author  = {Riveros Perez, Efrain and Avella-Molano, Bibiana},
  title   = {Learning from the machine: is diabetes in adults predicted by lifestyle variables? {A} retrospective predictive modelling study of {NHANES} 2007--2018},
  journal = {BMJ Open},
  year    = {2025},
  volume  = {15},
  number  = {3},
  pages   = {e096595},
  doi     = {10.1136/bmjopen-2024-096595}
}

@article{yang2026nhanes_highrisk,
author = {Yang, Xiaohua and Yao, Meiqi and Huang, Jia and Cheng, Zhuojing and Sun, Ting},
title = {Machine learning predicts diabetes risk in high-risk populations: analysis of {National Health and Nutrition Examination Survey} data},
journal = {Archives of Medical Science},
year = {2026},
volume = {22},
number = {2},
pages = {708--720},
doi = {10.5114/aoms/209547}
}

@article{hu2023nomogram,
  author  = {Hu, Yanhua and Han, Yong and Liu, Yufei and Cui, Yanan and Ni, Zhiping and Wei, Ling and Cao, Changchun and Hu, Haofei and He, Yongcheng},
  title   = {A nomogram model for predicting 5-year risk of prediabetes in {C}hinese adults},
  journal = {Scientific Reports},
  year    = {2023},
  volume  = {13},
  number  = {1},
  pages   = {22523},
  doi     = {10.1038/s41598-023-50122-3}
}

@article{almadhoun2025prediabetes,
author  = {Almadhoun, Mahmoud B. and Burhanuddin, M. A.},
title   = {Optimizing Feature Selection and Machine Learning Algorithms for Early Detection of Prediabetes Risk: Comparative Study},
journal = {JMIR Bioinformatics and Biotechnology},
year    = {2025},
volume  = {6},
pages   = {e70621},
doi     = {10.2196/70621}
}

@article{fan2026progmdd,
  author  = {Fan, Huaxin and Li, Zhendong and Yan, Ning and Liu, Hao},
  title   = {An interpretable progressive residual network for automated multiclass diabetes diagnosis},
  journal = {Scientific Reports},
  volume  = {16},
  number  = {1},
  pages   = {20468},
  year    = {2026},
  doi     = {10.1038/s41598-026-51603-x}
}

@article{huang2020tabtransformer,
author  = {Huang, Xin and Khetan, Ashish and Cvitkovic, Milan and Karnin, Zohar},
title   = {{TabTransformer}: Tabular Data Modeling Using Contextual Embeddings},
journal = {arXiv preprint arXiv:2012.06678},
year    = {2020},
doi     = {10.48550/arXiv.2012.06678}
}

@inproceedings{gorishniy2021fttransformer,
author    = {Gorishniy, Yury and Rubachev, Ivan and Khrulkov, Valentin and Babenko, Artem},
title     = {Revisiting Deep Learning Models for Tabular Data},
booktitle = {Advances in Neural Information Processing Systems},
volume    = {34},
pages     = {18932--18943},
year      = {2021}
}

@article{somepalli2021saint,
author  = {Somepalli, Gowthami and Goldblum, Micah and Schwarzschild, Avi and Bruss, C. Bayan and Goldstein, Tom},
title   = {{SAINT}: Improved Neural Networks for Tabular Data via Row Attention and Contrastive Pre-Training},
journal = {arXiv preprint arXiv:2106.01342},
year    = {2021},
doi     = {10.48550/arXiv.2106.01342}
}

@article{mukherjee2026xaiiot,
  author  = {Sunetra Mukherjee and Debashis De},
  title   = {{XAI-IoT} Based Real-Time Diabetes Prediction Using {TabTransformer}},
  journal = {Medicine in Novel Technology and Devices},
  volume  = {30},
  pages   = {100435},
  year    = {2026},
  doi     = {10.1016/j.medntd.2026.100435}
}

@inproceedings{kevin2025diabetes,
  author    = {{Kevin} and Naufal Krisna Thariq and Edy Irwansyah},
  title     = {Diabetes Risk Classification in Medical Data Using {TabTransformer}},
  booktitle = {2025 6th International Conference on Artificial Intelligence and Data Sciences ({AiDAS})},
  pages     = {50--55},
  year      = {2025},
  doi       = {10.1109/AIDAS67696.2025.11213882}
}

@inproceedings{khan2025early,
  author    = {Sulaiman Khan and Md. Rafiul Biswas and Zubair Shah},
  title     = {Early Prediction of Type 2 Diabetes Using Multimodal Data and Tabular Transformers},
  booktitle = {2025 3rd International Conference on Foundation and Large Language Models ({FLLM})},
  pages     = {592--599},
  year      = {2025},
  doi       = {10.1109/FLLM67465.2025.11391053}
}

@article{wang2025msft,
  author  = {Ning Wang and Minghui Wu and Wenchao Gu and Chenglong Dai and Zongru Shao and K. P. Subbalakshmi},
  title   = {{MSFT-Transformer}: A Multistage Fusion Tabular Transformer for Disease Prediction Using Metagenomic Data},
  journal = {Briefings in Bioinformatics},
  volume  = {26},
  number  = {3},
  pages   = {bbaf217},
  year    = {2025},
  doi     = {10.1093/bib/bbaf217}
}

@inproceedings{oba2021interpretable,
  author    = {Yuki Oba and Taro Tezuka and Masaru Sanuki and Yukiko Wagatsuma},
  title     = {Interpretable Prediction of Diabetes from Tabular Health Screening Records Using an Attentional Neural Network},
  booktitle = {2021 IEEE 8th International Conference on Data Science and Advanced Analytics ({DSAA})},
  pages     = {1--11},
  year      = {2021},
  doi       = {10.1109/DSAA53316.2021.9564151}
}

@article{knowler2002reduction,
  author  = {William C. Knowler and Elizabeth Barrett-Connor and Sarah E. Fowler and Richard F. Hamman and John M. Lachin and Elizabeth A. Walker and David M. Nathan and {Diabetes Prevention Program Research Group}},
  title   = {Reduction in the Incidence of Type 2 Diabetes with Lifestyle Intervention or Metformin},
  journal = {New England Journal of Medicine},
  volume  = {346},
  number  = {6},
  pages   = {393--403},
  year    = {2002},
  doi     = {10.1056/NEJMoa012512}
}

@article{doshivelez2017rigorous,
  author  = {Finale Doshi-Velez and Been Kim},
  title   = {Towards a Rigorous Science of Interpretable Machine Learning},
  journal = {arXiv preprint arXiv:1702.08608},
  year    = {2017},
  doi     = {10.48550/arXiv.1702.08608}
}

@article{lipton2018mythos,
  author  = {Zachary C. Lipton},
  title   = {The Mythos of Model Interpretability},
  journal = {Communications of the ACM},
  volume  = {61},
  number  = {10},
  pages   = {36--43},
  year    = {2018},
  doi     = {10.1145/3233231}
}

@inproceedings{kleinberg2017tradeoffs,
  author    = {Jon Kleinberg and Sendhil Mullainathan and Manish Raghavan},
  title     = {Inherent Trade-Offs in the Fair Determination of Risk Scores},
  booktitle = {8th Innovations in Theoretical Computer Science Conference ({ITCS} 2017)},
  series    = {Leibniz International Proceedings in Informatics ({LIPIcs})},
  volume    = {67},
  pages     = {43:1--43:23},
  year      = {2017},
  doi       = {10.4230/LIPIcs.ITCS.2017.43}
}

@article{he2009imbalanced,
  author  = {Haibo He and Edwardo A. Garcia},
  title   = {Learning from Imbalanced Data},
  journal = {IEEE Transactions on Knowledge and Data Engineering},
  volume  = {21},
  number  = {9},
  pages   = {1263--1284},
  year    = {2009},
  doi     = {10.1109/TKDE.2008.239}
}

@article{mcinnes2020umapuniformmanifoldapproximation,
  author  = {McInnes, Leland and Healy, John and Melville, James},
  title   = {{UMAP}: Uniform Manifold Approximation and Projection for Dimension Reduction},
  journal = {arXiv preprint arXiv:1802.03426},
  year    = {2020},
  doi     = {10.48550/arXiv.1802.03426}
}

@article{ada2026,
  author  = {{American Diabetes Association Professional Practice Committee for Diabetes}},
  title   = {{2. Diagnosis and Classification of Diabetes: Standards of Care in Diabetes---2026}},
  journal = {Diabetes Care},
  volume  = {49},
  number  = {Supplement 1},
  pages   = {S27--S49},
  year    = {2026},
  doi     = {10.2337/dc26-S002}
}






\end{document}